%% file: main.tex
\documentclass[10pt,twocolumn,letterpaper]{article}

\usepackage[pagenumbers]{iccv} 

\input{preamble}

\definecolor{iccvblue}{rgb}{0.21,0.49,0.74}
\usepackage[pagebackref,breaklinks,colorlinks,allcolors=iccvblue]{hyperref}

\usepackage{booktabs} 
\usepackage{multirow}  
\usepackage{graphicx}
\usepackage{adjustbox}
\usepackage{subcaption}
\usepackage{mdframed}
\usepackage[most]{tcolorbox}

\newcommand{\name}{IConMark}

\def\paperID{14712} 
\def\confName{ICCV}
\def\confYear{2025}

\title{IConMark: Robust Interpretable Concept-Based Watermark For AI Images}

\author{Vinu Sankar Sadasivan \qquad Mehrdad Saberi \qquad Soheil Feizi \\
Department of Computer Science\\
University of Maryland, College Park, USA\\
\texttt{\{vinu, msaberi, sfeizi\}@cs.umd.edu} 
}

\begin{document}
\maketitle

\begin{abstract}
With the rapid rise of generative AI and synthetic media, distinguishing AI-generated images from real ones has become crucial in safeguarding against misinformation and ensuring digital authenticity. 
Traditional watermarking techniques have shown vulnerabilities to adversarial attacks, undermining their effectiveness in the presence of attackers.
We propose \name, a novel in-generation robust semantic watermarking method that embeds interpretable concepts into AI-generated images, as a first step toward interpretable watermarking. 
Unlike traditional methods, which rely on adding noise or perturbations to AI-generated images, \name ~incorporates meaningful semantic attributes, making it interpretable to humans and hence, resilient to adversarial manipulation.
This method is not only robust against various image augmentations but also human-readable, enabling manual verification of watermarks. 
We demonstrate a detailed evaluation of IConMark’s effectiveness, demonstrating its superiority in terms of detection accuracy and maintaining image quality. 
Moreover, \name ~can be combined with existing watermarking techniques to further enhance and complement its robustness. 
We introduce \name+SS and \name+TM, hybrid approaches combining \name ~with StegaStamp and TrustMark, respectively, to further bolster robustness against multiple types of image manipulations.
Our base watermarking technique (\name) and its variants (+TM and +SS) achieve 10.8\%, 14.5\%, and 15.9\% higher mean area under the receiver operating characteristic curve (AUROC) scores for watermark detection, respectively, compared to the best baseline on various datasets.
\end{abstract}

\input{1_intro}

\input{2_related}
\input{3_0_method}
\input{3_1_combined}
\input{4_results}
\input{5_conclusion}

\section*{Acknowledgements}
The authors thank Matthijs Douze, Jakob Verbeek, and Pierre Fernandez for their support and mentorship throughout the project.
This project was supported in part by a grant from an NSF CAREER AWARD 1942230, ONR YIP award N00014-22-1-2271, ARO’s Early Career Program Award 310902-00001, Army Grant No. W911NF2120076, the NSF award CCF2212458, NSF Award No. 2229885 (NSF Institute for Trustworthy AI in Law and Society, TRAILS), a MURI grant 14262683, an award from meta 314593-00001 and an award from Capital One.
\bibliography{main}
\bibliographystyle{ieeenat_fullname}

\input{6_app}


\end{document}

%% file: preamble.tex
\pdfpageattr{/Group <</S /Transparency /I true /CS /DeviceRGB>>}

%% file: 1_intro.tex
\begin{figure*}[ht]
    \centering
    \includegraphics[width=0.99\linewidth]{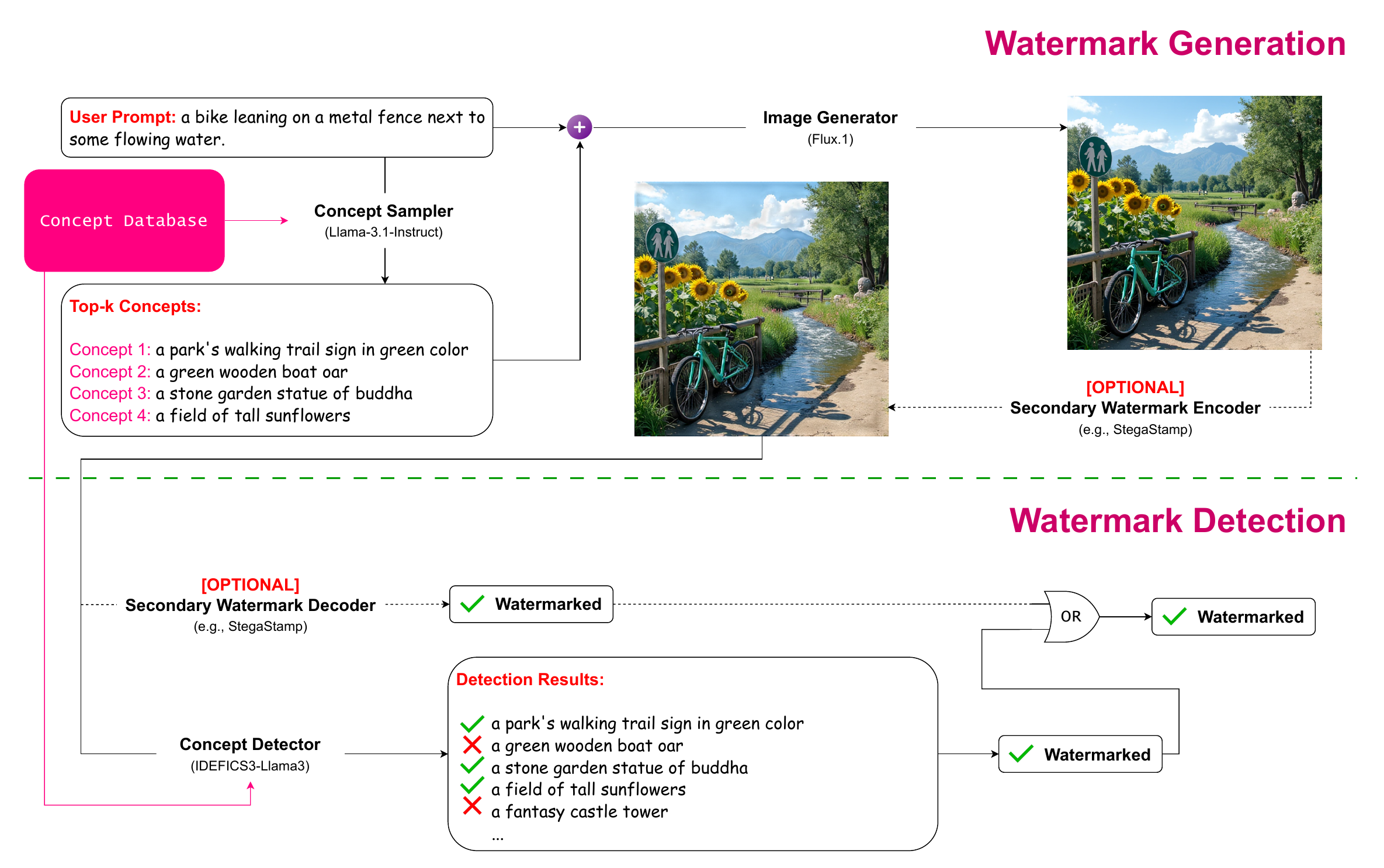}
    \caption{Watermark generation (top) and detection (bottom) pipeline for \name{} (and variants). Various concepts are generated and stored in a concept database (Section~\ref{sec:concept_generation}). User prompts are augmented with sampled concepts to generate watermarked images with interpretable concepts (Section~\ref{sec:watermark_generation}. A visual language model is used to query for the presence of database concepts in a candidate image for watermark detection (Section~\ref{sec:watermark_detection}). Additional watermarking can be combined with \name{} to further bolster its robustness (Section~\ref{sec:combine}).}
    \label{fig:main_fig_method}
\end{figure*}

\section{Introduction}
With the rapid advancements in generative AI, distinguishing between AI-generated and real images has become a critical challenge. 
The proliferation of deepfake technologies and synthetic media has raised concerns about misinformation, copyright infringement, and digital authentication \citep{helmus2022artificial, christodorescu2024securing}. 
Traditional watermarking techniques, which add imperceptible noise or frequency domain modifications to images, have been widely used to establish provenance and protect intellectual property \citep{dwtdct, stegastamp, trustmark, treering, wam, stablesignature, saberi2024drew}. 
However, recent research has demonstrated the vulnerability of existing watermarking methods to adversarial attacks, including diffusion purification and model substitution adversarial attacks, which can remove or spoof watermarks with minimal image alterations \citep{saberi2023robustness}.

Watermarking methods can broadly be categorized into post-hoc and in-generation approaches. 
Post-hoc watermarks modify an image after the generation, embedding signals through additive noise or frequency-space perturbations \citep{stegastamp, stablesignature, dwtdct, trustmark}. 
In contrast, in-generation watermarks embed information during the image generation process, modifying the distribution of generated samples rather than making explicit post-hoc changes \citep{treering, prc, yang2024gaussian}. 
While in-generation watermarking methods tend to be theoretically more robust, they are still susceptible to attacks, particularly adversarial strategies that attempt to remove or obfuscate the watermark \citep{saberi2023robustness}.

In this work, we introduce
a robust \underline{I}nterpretable \underline{Con}cept-based Water\underline{mark} (\name), a novel in-generation semantic watermarking approach that embeds interpretable concepts into AI-generated images, as a first step toward interpretable AI watermarking.
Unlike traditional watermarks, which primarily rely on additive noise, our method integrates meaningful semantic attributes, making it robust against various image perturbations and adversarial purification attacks.
Since these concepts are interpretable to humans, we can also manually check if an image generated with \name ~is watermarked. 
This novelty makes \name ~interpretable and ~robust to adversarial attacks, potentially making it a strong technique for manual image forensics with the help of human experts.
\name ~can also automate detection using a visual language model, which queries the presence of these embedded concepts.
This ensures both machine verifiability and human interpretability. 
Furthermore, we demonstrate that
\name ~can be effectively combined with any existing watermarking approach since it only perturbs the input prompt given to the image generation model.
We demonstrate that this ability of our \name ~to complement existing watermarking approaches can result in strong robust image watermarking techniques. 

Our contributions are as follows:

\begin{itemize}
\item We propose
\name ~(Section~\ref{sec:method}), a novel semantic watermarking method that embeds interpretable concepts rather than additive noise, improving the robustness and making AI image watermark interpretable for the first time.

\item \name ~by design can be effectively combined with any existing watermarking technique to further enhance robustness against image augmentations or attacks (Section~\ref{sec:combine}).

\item We show that our method is resistant to various image augmentation attacks, including diffusion purification attacks (Section~\ref{sec:experiments}). \name ~being interpretable makes it easier to be detected by human experts, hence making our method resilient to adversarial attacks.

\item We evaluate the quality of generation with
\name, showing that it maintains high visual and generation quality while ensuring watermark detection integrity.
\end{itemize}

%% file: 2_related.tex
\section{Related Works}

Traditional and deep learning-based watermarking has long been a crucial tool for copyright protection, content authenticity, and AI-generated media detection \citep{honsinger2002digital, swanson1998multimedia}. 
Classical techniques embed signals in the spatial or frequency domain, leveraging transformations such as Discrete Cosine Transform (DCT) and Discrete Wavelet Transform (DWT) \citep{al2007combined, dwtdct}. 
Deep learning-based watermarking methods, such as StegaStamp \citep{stegastamp}, StableSignature \citep{stablesignature}, TrustMark \citep{trustmark}, and Watermark Anything Model \citep{wam} have further improved robustness by training neural networks to encode and extract watermarks from images. 
However, these methods remain susceptible to sophisticated removal techniques, including adversarial attacks and diffusion purification \citep{saberi2023robustness}.

\cite{saberi2023robustness} demonstrated that diffusion purification attacks, which leverage denoising diffusion models, can effectively remove low-perturbation watermarks by reconstructing images with minimal modifications. 
Similarly, WAVES \citep{waves}, a benchmarking framework, revealed that watermark detection accuracy significantly degrades under adversarial and regeneration attacks, highlighting the need for more robust watermarking solutions. 
Additionally, high-perturbation watermarking methods, such as TreeRing \citep{treering}, have been shown to be more resistant to removal but are still vulnerable to adversarial model substitution attacks \citep{saberi2023robustness} and simple image averaging attacks \citep{yang2024can}.

Furthermore, \cite{saberi2023robustness, sadasivan2023can} show that watermarking systems are not only vulnerable to removal but also to spoofing attacks, where real images are falsely classified as watermarked, leading to false attributions and reputational risks \citep{saberi2023robustness}.
Adversarial perturbations designed to mimic the statistical properties of watermarked images have been shown to effectively fool detection systems, raising questions about the reliability of current watermarking schemes.
Our work builds on these findings by introducing an approach that is inherently more interpretable and robust to such attacks.

The concept of semantic watermarking, which embeds meaningful and interpretable information into images, has been unexplored. 
Existing methods primarily focus on imperceptibility and robustness but often lack interpretability. 
Our work builds on these foundations by integrating interpretable concepts into AI-generated images, ensuring both human and machine verifiability while maintaining robustness against adversarial attacks for the first time.
Our approach aligns with recent trends in AI provenance tracking and content authentication, presenting a compelling proof-of-concept for the future of interpretable watermarking technologies.

%% file: 3_0_method.tex
\section{\name: Interpretable Watermarking}
\label{sec:method}

In this section, we describe our proposed method \name ~in detail (see Figure~\ref{fig:main_fig_method}).
For every user prompt for image generation, \name ~samples related concepts from a private database.
\name ~augments the user prompt with the sampled concepts and uses this augmented user prompt to query the image generator.
The AI-image generator performs in-generation watermarking by adding these interpretable syntactical signatures or concepts to the generated images.
At detection time for a candidate image, \name ~checks for the presence of various concepts from the private concept database with the aid of a visual language model.
If the candidate image has more than a threshold number of concepts from the private concept database, it is classified as watermarked. Below, we describe various steps for the proposed watermarking pipeline.

\subsection{Initialization: Concept Database Generation}\label{sec:concept_generation}
We prompt ChatGPT \citep{OpenAI2023ChatGPT} to generate multiple image concepts that can be added to an image. 
We instruct the model to generate concepts describing simple objects with a unique detail. 
For example, `` brass table lamp'' or ``a metal blue street sign''.
We also instruct the model to generate concepts that can occur in various settings such as indoors, nature, sky, forest, streets, etc.
We then manually craft a diverse database $\mathcal{D}$ with $N$ concepts.
Note that the concept database generation needs to be performed pre-hoc only once for our setup.
This $\mathcal{D}$ can be later used to automatically generate or detect any new \name ~watermarked image without any manual intervention as shown in the following subsections.

\subsection{Watermarked Image Generation}\label{sec:watermark_generation}
For every user prompt $p$, \name ~samples top-$k$ related concepts from the private concept database $\mathcal{D} = \{c_1, c_2,..., c_N\}$.
\name ~prompts Llama-3.1-8B-Instruct model, $\mathcal{L}$, to sample the related concepts.
Llama gets the database $\mathcal{D}$, the user prompt $p$, and the number of concepts $k$ to be sampled as inputs using a custom system prompt template to sample concepts $c_p^1, c_p^2, ..., c_p^k \in \mathcal{D}$. 
Here is the custom prompt template for $\mathcal{L}$:

\begin{tcolorbox}[breakable, enhanced]
\textbf{System prompt}: Here is a database of $N$ concepts: \textbackslash n \textbackslash n
$c_1$ \textbackslash n $c_2$ \textbackslash n \ldots \textbackslash n $c_N$ \textbackslash n \textbackslash n

\noindent In an image of `$p$', what are the top $k$ related concepts from this database that can very likely occur in the background of this image? Consider only concepts that are related to this given image. For example, an image of a lion cannot have a basketball or a table in the background, whereas an image of a bird can have a tree or a mountain in the background. The concepts should be ONLY from the database of concepts given above. You should NOT generate new concepts.

\textbf{User}: Print each of the $k$ related concepts verbatim between \textless a \textgreater and \textless /a \textgreater.
\end{tcolorbox}

\name ~then uses a prompt template to generate an augmented user prompt $\tilde{p}_k$.
\name ~passes $\tilde{p}_k$ to the image generator $\mathcal{G}$ to generate a watermarked image with the syntactical concept-based signatures.
Here is the augmented prompt template for $\tilde{p}_k$:

\begin{tcolorbox}[breakable, enhanced]
$p$ in the foreground. add following: \textbackslash n
$c_p^1$ \textbackslash n $c_p^2$ \textbackslash n \ldots \textbackslash n $c_p^k$. \textbackslash n \textbackslash n sharp, detailed.
\end{tcolorbox}

\subsection{Watermark detection}\label{sec:watermark_detection}
For a candidate image $x$, \name ~has access to the private concept database $\mathcal{D}$ and $\mathcal{V}$, a visual language model, IDEFICS3-8B-Llama3.
\name ~prompts $\mathcal{V}$ to check for the presence of each of the $N$ concepts in $\mathcal{D}$ given the image $x$ using a custom prompt template.
Here is the prompt template for prompting $\mathcal{V}$ to detect the presence of a concept $c_i \in \mathcal{D}$:

\begin{tcolorbox}[breakable, enhanced]
\textbf{Image input}: $x$

\noindent\textbf{Text input}: Print yes or no. Is there something like `$c_i$'?
\end{tcolorbox}

The detection score is the number of objects in $\mathcal{D}$ that were detected in $x$ by $\mathcal{V}$.
If the detection score is greater than a threshold $\tau$, the candidate image $x$ is classified as watermarked. 
Else, the image is labeled as non-watermarked.

%% file: 3_1_combined.tex
\begin{figure*}[h]
    \centering
    \includegraphics[width=0.83\linewidth]{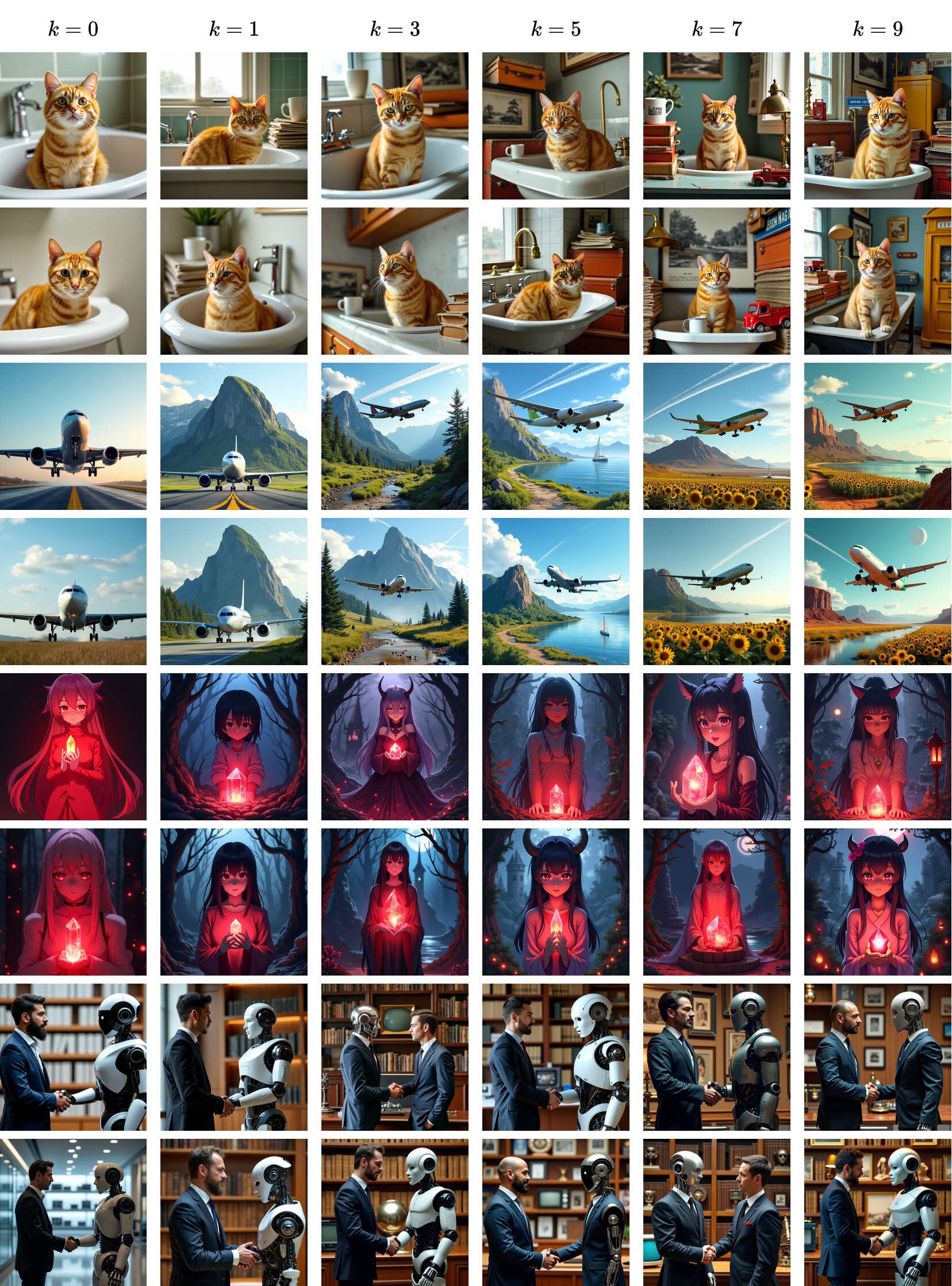}
    \caption{Illustration of images generated by the Flux model and \name{} for different values of $k$. Each prompt has two rows of images to highlight the changes in distribution of generated images with ($k>0$) and without ($k=0$) \name{} watermarking. From top to bottom, the prompts are “A yellow striped cat sitting in a bathroom sink” and “An airplane with its landing wheels out landing” (from MS-COCO), followed by “An anime character illuminated by a red crystal against a dark backdrop” and “A man shakes hands with a robot, both wearing business suits in an office or library” (from OIP).}
    \label{fig:main_fig_dists}
\end{figure*}

\begin{figure*}[t]
    \centering
    \begin{minipage}{0.48\textwidth}
        \centering
        \includegraphics[width=\textwidth]{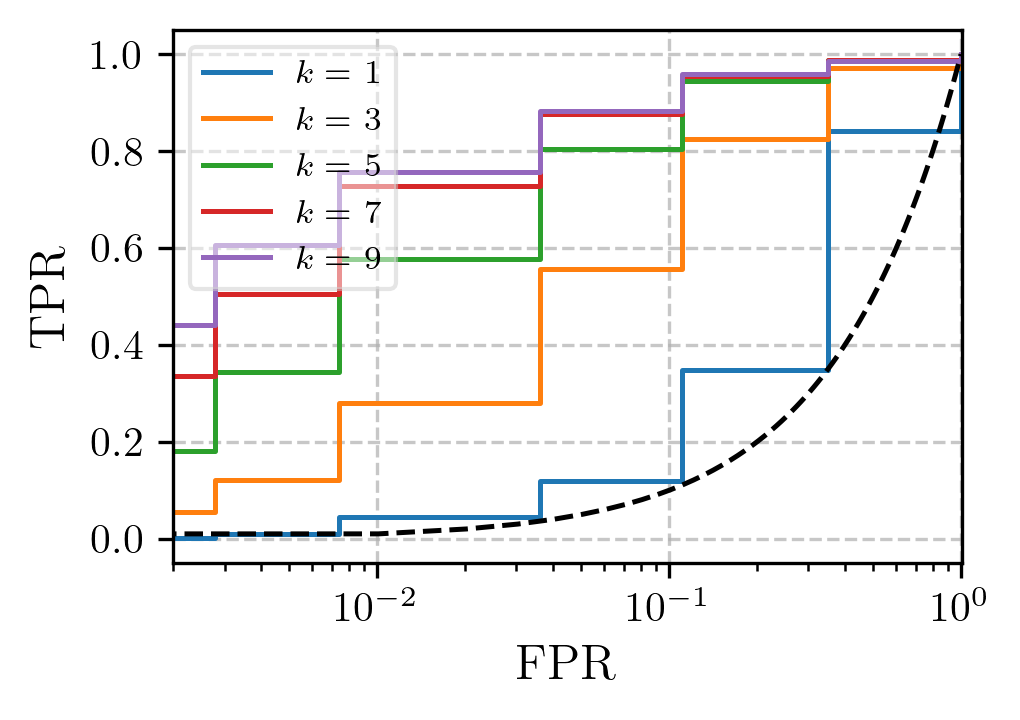}
        \label{fig:topk_ablation}
    \end{minipage}%
    \begin{minipage}{0.48\textwidth}
        \centering
        \includegraphics[width=\textwidth]{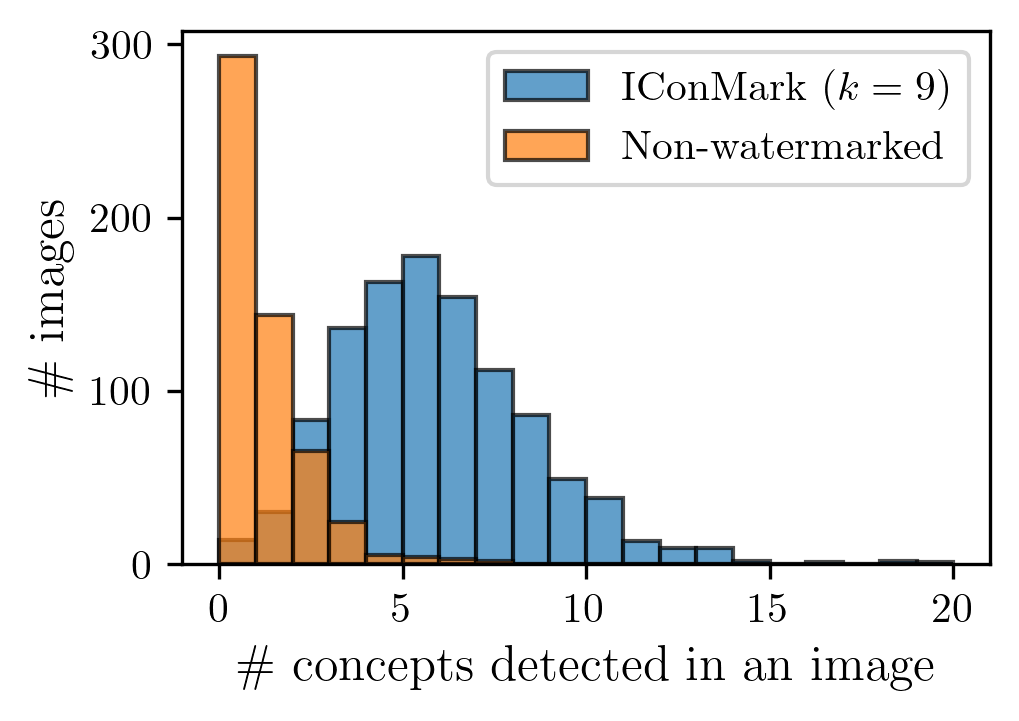}
        \label{fig:concepts_hist}
    \end{minipage}
    \vspace{-0.5cm}
    \caption{(Left) ROC curves for ablations of $k$ for our proposed method \name. The black dashed lines indicate the ROC curve of a random detector. (Right) Number of concepts detected in watermarked (\name ~$k=9$) and non-watermarked images by the IDEFICS3 visual language model.}
    \label{fig:ablations}
\end{figure*}

\section{\name(+SS and +TM): Harnessing the Combinatorial Strength of \name}
\label{sec:combine}

\name{} can be combined with any existing watermarking method since \name{} only augments the input prompt.
In this section, we harness the combined strength of our method \name with StegaStamp \citep{stegastamp} and TrustMark \citep{trustmark}. 
\cite{waves} endorse StegaStamp for its resilience to various image manipulations.
StegaStamp stood out as a robust watermark among other popular techniques such as TreeRing \citep{treering} and StableSignature \citep{stablesignature} in the WAVES benchmarking \citep{waves}.

However, unlike \name ~these prior watermarking techniques lack interpretability and are not robust to adversarial attacks \citep{saberi2023robustness}.
Since \name ~can be complementary to any existing watermarking technique, we propose to combine the powers of both StegaStamp (or TrustMark) and our method \name ~to bolster detection robustness. 
We name this method \name+SS (or \name+TM, respectively).

\name+SS (or +TM) generates \name ~watermarked images ~and then applies post-hoc watermarking using StegaStamp (or TrustMark, respectively).
At detection time, \name+SS (or +TM) labels an image as watermarked if either the \name ~or StegaStamp (or TrustMark) detectors label it as watermarked. 
If both the detectors label the image as non-watermarked, only then the image is labeled as non-watermarked by \name+SS (or +TM). 
This makes \name+SS (or +TM) robust to all the attacks that both StegaStamp (or TrustMark) and \name ~are resilient to.

%% file: 4_results.tex
\section{Experiments}
\label{sec:experiments}

\begin{table}[t]
\centering
\begin{adjustbox}{max width=\linewidth}
\begin{tabular}{c | cccc}
\toprule
$\mathbf{k}$ & \textbf{AUC (\%)} & \textbf{Accuracy (\%)} & \textbf{T@5\%F (\%)} & \textbf{T@1\%F (\%)} \\
\midrule \midrule
\multicolumn{5}{c}{\textbf{MS-COCO}} \\
\midrule
1 & 76.05 & 74.58 & 11.85 & 4.44 \\
3 & 92.04 & 85.65 & 55.65 & 28.06 \\
5 & 96.47 & 91.67 & 80.46 & 57.69 \\
7 & 97.31 & 92.18 & 87.69 & 72.78 \\
9 & 97.46 & 92.41 & 88.24 & 75.65 \\
\midrule
\multicolumn{5}{c}{\textbf{OIP}} \\
\midrule
7 & 87.95 & 81.23 & 48.28 & 13.30 \\
9 & 87.92 & 81.74 & 50.41 & 18.34 \\
\bottomrule
\end{tabular}
\end{adjustbox}
\caption{Detection performance metrics for different values of $k$ for MS-COCO and OIP datasets.}
\label{tab:detection_metrics}
\end{table}

\begin{table}[t]
\centering
\begin{adjustbox}{max width=\linewidth}
\begin{tabular}{c | cccc}
\toprule
$\mathbf{k}$ & \textbf{Clip Score} $\uparrow$ &  \textbf{Diversity} $\uparrow$ & \textbf{Ratings} $\uparrow$ & \textbf{Artifacts} $\downarrow$  \\
\midrule \midrule
\multicolumn{5}{c}{\textbf{MS-COCO}} \\
\midrule
1 & 0.026 $\pm$ 0.030 & 0.285 $\pm$ 0.014 &  6.073 $\pm$ 1.016 & 2.156 $\pm$ 0.631 \\
3 & 0.029 $\pm$ 0.031 & 0.307 $\pm$ 0.018 &  6.069 $\pm$ 1.042 & 2.164 $\pm$ 0.633 \\
5 & 0.031 $\pm$ 0.032 & 0.325 $\pm$ 0.019 &  6.123 $\pm$ 0.998 & 2.139 $\pm$ 0.622 \\
7 & 0.032 $\pm$ 0.032 & 0.342 $\pm$ 0.021 &  6.250 $\pm$ 1.028 & 2.102 $\pm$ 0.613 \\
9 & 0.033 $\pm$ 0.032 & 0.349 $\pm$ 0.022 &  6.282 $\pm$ 1.084 & 2.100 $\pm$ 0.632 \\
\midrule
\multicolumn{5}{c}{\textbf{OIP}} \\
\midrule
7 & 0.064 $\pm$ 0.039 & 0.332 $\pm$ 0.019 &  7.284 $\pm$ 1.187 & 1.723 $\pm$ 0.591 \\
9 & 0.066 $\pm$ 0.039 & 0.333 $\pm$ 0.019 &  7.321 $\pm$ 1.196 & 1.715 $\pm$ 0.600 \\
\bottomrule
\end{tabular}
\end{adjustbox}
\caption{Image generation metrics for different values of $k$ for MS-COCO and OIP datasets.}
\label{tab:generation_metrics}
\end{table}


\begin{table*}[t]
    \centering
    \begin{adjustbox}{max width=\linewidth}
    \begin{tabular}{c | ccc | ccc}
        \toprule
        \textbf{Metrics} & \textbf{DWTDCT} & \textbf{TrustMark} & \textbf{StegaStamp} & \textbf{IConMark} & \textbf{\name+TM} & \textbf{\name+SS} \\
        \midrule \midrule
        \multicolumn{7}{c}{\textbf{No augmentations}} \\
        \midrule
        AUC & 100.00 (99.69) & 100.00 (100.00) & 100.00 (100.00) & 97.46 (87.92) & 100.00 (100.00) & 100.00 (100.00)  \\
        Accuracy & 100.00 (99.00) & 100.00 (100.00) & 100.00 (100.00) & 92.41 (81.74) & 100.00 (100.00) & 100.00 (100.00)  \\
        T@5\%F & 100.00 (98.27) & 100.00 (100.00) & 100.00 (100.00) & 88.24 (50.41) & 100.00 (100.00) & 100.00 (100.00)  \\
        T@1\%F & 100.00 (98.27) & 100.00 (100.00) & 100.00 (100.00) & 75.65 (18.34) & 100.00 (100.00) & 100.00 (100.00)  \\
        \midrule
        \multicolumn{7}{c}{\textbf{Affine augmentations}} \\
        \midrule
        AUC & 73.52 (73.18) & 70.00 (68.63) & 55.41 (52.79) & 96.27 (86.30) & \textbf{97.40} (\textbf{92.40}) & \underline{96.32} (\underline{86.41}) \\
        Accuracy & 70.32 (69.73) & 68.89 (68.45) & 54.58 (53.05) & 90.65 (80.23) & \textbf{93.43} (\textbf{85.23}) & \underline{90.65} (\underline{80.45}) \\
        T@5\%F & 3.43 (7.82) & 42.31 (40.18) & 4.72 (3.64) & 84.07 (31.82) & \textbf{90.37} (\textbf{60.91}) & \underline{84.63} (\underline{32.27}) \\
        T@1\%F & 1.20 (0.55) & 38.15 (37.18) & 1.02 (0.64) & 51.11 (13.18) & \textbf{69.07} (\textbf{48.64}) & \underline{51.11} (\underline{13.64}) \\
        \midrule
        \multicolumn{7}{c}{\textbf{Regen augmentations}} \\
        \midrule
        AUC & 60.29 (56.04) & 63.07 (59.31) & \underline{96.54} (\underline{95.81}) & 96.14 (86.34) & 96.21 (86.36) & \textbf{99.05} (\textbf{97.46}) \\
        Accuracy & 56.90 (53.95) & 59.35 (56.59) & 90.09 (\underline{89.18}) & 91.39 (78.86) & \underline{91.48} (79.09) & \textbf{95.46} (\textbf{92.27}) \\
        T@5\%F & 12.87 (11.73) & 11.30 (8.91) & \underline{82.50} (\underline{73.55}) & 74.63 (44.09) & 76.11 (49.09) & \textbf{95.19} (\textbf{81.36}) \\
        T@1\%F & 3.24 (4.91) & 2.96 (1.73) & \underline{60.46} (\underline{62.00}) & 59.26 (13.64) & 59.26 (17.27) & \textbf{79.44} (\textbf{69.09}) \\
        \midrule
        \multicolumn{7}{c}{\textbf{Valuemetric augmentations}} \\
        \midrule
        AUC & 45.50 (53.85) & 85.75 (93.37) & \underline{99.83} (\underline{99.33}) & 93.40 (83.43) & 96.66 (95.51) & \textbf{99.93} (\textbf{99.77}) \\
        Accuracy & 51.76 (53.45) & 78.75 (89.14) & \underline{98.98} (\underline{98.18}) & 87.50 (75.68) & 91.57 (90.00) & \textbf{99.35} (\textbf{98.64}) \\
        T@5\%F & 5.74 (7.18) & 56.20 (83.00) & \underline{99.26} (\underline{97.73}) & 67.04 (38.64) & 84.81 (85.00) & \textbf{99.44} (\textbf{98.64}) \\
        T@1\%F & 0.83 (2.45) & 41.94 (77.36) & \underline{98.52} (\underline{96.73}) & 32.22 (12.27) & 54.81 (77.73) & \textbf{99.26} (\textbf{97.73}) \\
        \midrule
        \multicolumn{7}{c}{\textbf{Warp augmentations}} \\
        \midrule
        AUC & 50.74 (51.05) & 51.02 (51.50) & 49.52 (52.41) & 95.67 (86.30) & \textbf{95.71} (\textbf{86.35}) & \underline{95.68} (\underline{86.31}) \\
        Accuracy & 51.44 (51.82) & 51.25 (51.86) & 50.60 (\underline{52.77}) & \underline{90.97} (\textbf{80.23}) & \textbf{91.20} (\textbf{80.23}) & \underline{90.97} (\textbf{80.23}) \\
        T@5\%F & 0.00 (0.09) & 6.48 (6.27) & 2.69 (5.45) & \underline{72.69} (\underline{31.82}) & \textbf{74.07} (\textbf{32.73}) & \underline{72.69} (\underline{31.82}) \\
        T@1\%F & 0.00 (0.00) & 1.94 (0.73) & 0.65 (\underline{1.00}) & \underline{41.67} (\textbf{13.18}) & \textbf{42.59} (\textbf{13.18}) & \underline{41.67} (\textbf{13.18}) \\
        \midrule \midrule
        \textbf{Average AUC} & 66.01 (66.76) & 73.97 (74.56) & 80.26 (80.07) & 95.79 (86.05) & \underline{97.20} (\underline{92.12}) & \textbf{98.12} (\textbf{93.99}) \\
        \bottomrule
    \end{tabular}
    \end{adjustbox}
    \caption{Comparison of \name{} and its variants to various baseline methods with and without various image augmentation attacks. The displayed numbers represent detection metrics on the MS-COCO dataset, while the numbers in parentheses correspond to the OIP dataset. \textbf{Bold} values indicate the best detector(s) in each setting, and \underline{underlined} values denote the second-best detector(s).}
    \label{tab:augmentation_comparison}
\end{table*}

\begin{figure*}[t]
    \centering
    \begin{minipage}{0.48\textwidth}
        \centering
        \includegraphics[width=\textwidth]{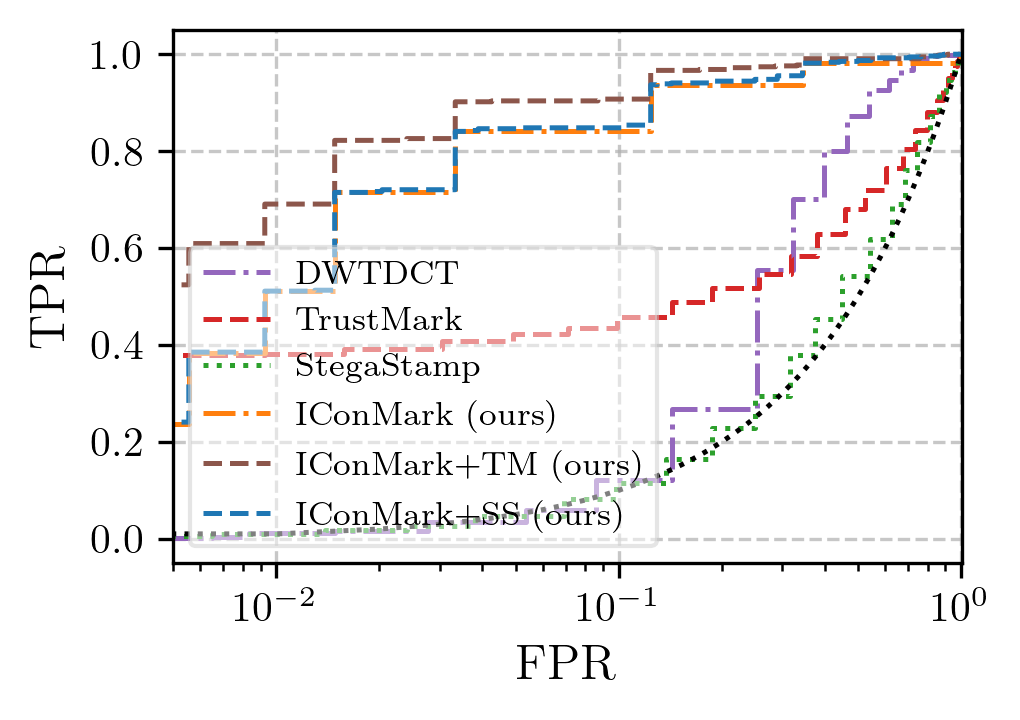}
        \subcaption{Affine}\label{fig:geometric_aug}
    \end{minipage}%
    \begin{minipage}{0.48\textwidth}
        \centering
        \includegraphics[width=\textwidth]{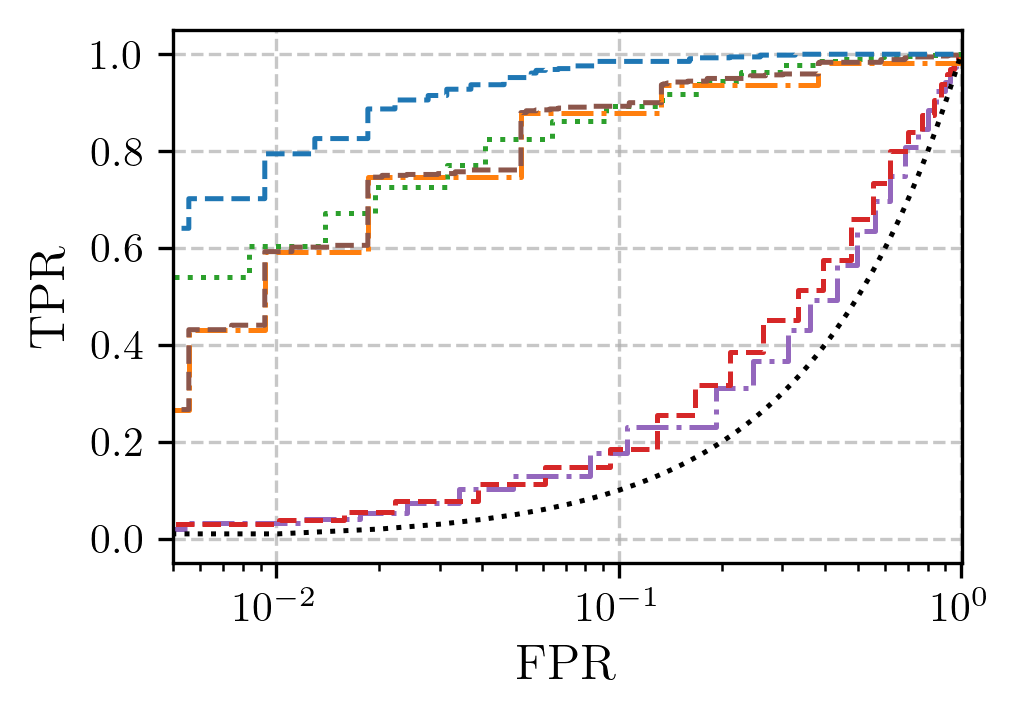}
        \subcaption{Regen}\label{fig:regen_aug}
    \end{minipage}
    \begin{minipage}{0.48\textwidth}
        \centering
        \includegraphics[width=\textwidth]{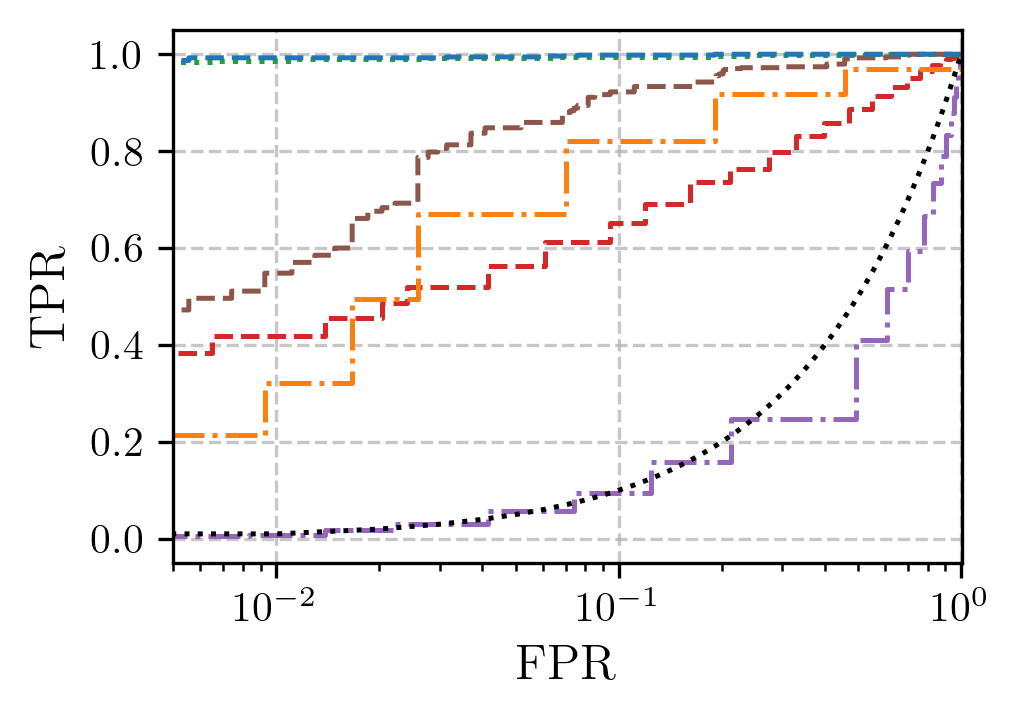}
        \subcaption{Valuemetric}\label{fig:valuemetric_aug}
    \end{minipage}%
    \begin{minipage}{0.48\textwidth}
        \centering
        \includegraphics[width=\textwidth]{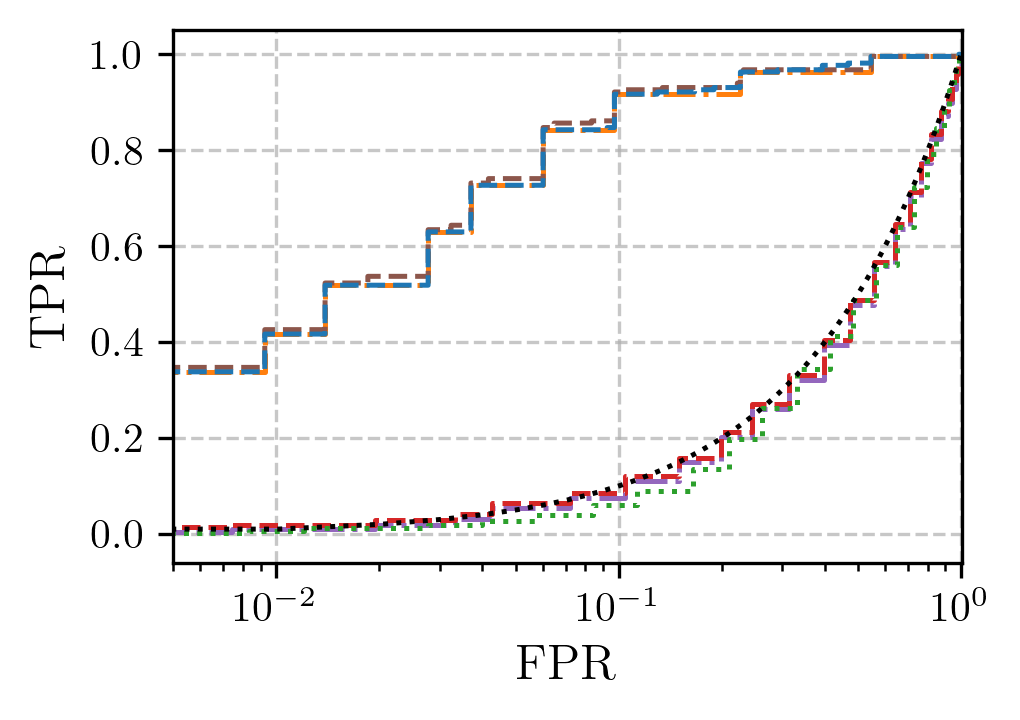}
        \subcaption{Warp}\label{fig:warp_aug}
    \end{minipage}
    
    \caption{ROC curves of various watermarking techniques in the presence of various image augmentations with the MS-COCO dataset.
    Black dotted curves indicate the performance of a random detector.
    As shown, our \name{} variants perform the best consistently.}
    \label{fig:robustness}
\end{figure*}

In this section, we provide the experiments to demonstrate the effectiveness of our watermarks.
Section~\ref{sec:setup} provides the experimental setup of our work describing the baselines, dataset, models, and metrics used.
In Section~\ref{sec:detection}, we show the performance of \name ~and its variants \name+SS (and +TM) when compared to other baselines. 
We also perform ablation studies over hyperparameters used for \name ~and quantify the effect of changes in image quality due to watermarking.
See Figure~\ref{fig:main_fig_dists} for images generated via \name ~with different values of $k$.
Section~\ref{sec:robustness} shows the experiments testing various watermarks in the presence of different image manipulation techniques to study their robustness to such modifications.
Our experiments show that \name, being the only interpretable watermark, is also the clear winner maintaining detection performance before and after all the image augmentation attacks when compared to other baselines.
We use this to our advantage to complement the power of \name ~with StegaStamp (and TrustMark) to demonstrate the robustness of \name+SS (and +TM).
In our experiments, \name{} and its variants (+TM and +SS) achieve 10.8\%, 14.5\%, and 15.9\%
higher mean AUROC scores for watermark detection, respectively, compared to the StegaStamp on various datasets.

\subsection{Experimental Settings}
\label{sec:setup}

\textbf{Baselines.} 
We compare \name{} to several recent watermarking baselines, including StegaStamp \citep{stegastamp}, TrustMark \citep{trustmark}, and DwtDctSVD \citep{dwtdct}. All methods employ 100-bit binary watermark keys and encode the watermark by computing additive noise patterns that are applied to the images. 

\textbf{Dataset and Models.}
We use 108 captions from the MS-COCO dataset \citep{coco} and 110 captions from a harder Open Image Preferences (OIP) dataset\footnote{\url{https://huggingface.co/datasets/data-is-better-together/open-image-preferences-v1}} to generate AI images using the FLUX.1-dev model \citep{flux2024}. Throughout the paper, we refer to the image generation model $\mathcal{G}$ as Flux.
For sampling the top related image concepts for \name, we use the Llama-3.1-8B-Instruct model \citep{llama3}.
Throughout the paper, we refer to the language model $\mathcal{L}$ as Llama.
We use IDEFICS3-8B-Llama3 \citep{idefics3} as the visual language model for measuring \name ~detection score.
Throughout the paper, we refer to the visual language model $\mathcal{V}$ as Idefics.
In all our experiments, we generate 10 different images per prompt with $\mathcal{G}$.
For example, we use 1080 non-watermarked images and 1080 watermarked images for our main experiments, totaling 2160 images in our evaluation with MS-COCO captions (and 2200 images with OIP).
For all our detection robustness experiments with MS-COCO, we halve our dataset size to 1080 images. 
For \name, we use a concept database of size $N=100$ and, by default, sample $k=9$ concept to augment the user prompt. The list of all the concepts is provided in the Appendix.

\textbf{Detection Metrics.}
We plot the Receiver Operating Characteristic (ROC) curves or the True Positive Rates (TPR) vs. False Positive Rates (FPR) for the detection tasks.
We measure the area under the ROC curves (AUROC) and the accuracy of the detection methods.
We also measure TPR at 5\% FPR (T@5\%F) and TPR at 1\% FPR (T@1\%F).
It is quite straightforward to measure these detection metrics for any watermarking detector that provides a continuous range of detection scores.
However, for \name+SS (or +TM), we do not have an explicit detection score since it performs detection conditioned on the output of both \name ~and StegaStamp (or TrustMark) detectors.
Therefore, we measure the TP values of \name+SS (or +TM) at different FP values by varying their detection thresholds.
This gives us the TPR and FPR for \name+SS (or +TM).
To obtain the ROC curves, we then sample the Pareto-optimal detection threshold pairs of \name ~and StegaStamp (or TrustMark), ensuring that no selected point in the curve has a higher FPR for the same or lower TPR.
After obtaining the ROC curve in this manner, we measure the detection metrics for \name+SS (or +TM).

\textbf{Generation quality metrics.}
We use Clip Score \citep{clip, hessel2021clipscore}, Ratings (aesthetic score), Artifacts \citep{aesthetic}, and Diversity \citep{astolfi2024consistency} metrics to measure the image generation qualities.
Clip score measures the similarity of the generated image to the original user prompt text with respect to the CLIP model \citep{clip}.
Ratings and Artifacts measure the changes in aesthetic and artifact features of images using a fine-tuned  CLIP image reward model.
Diversity quantifies the capability of an image generator to produce diverse images for a prompt.

\subsection{Watermark Detection And Image Quality}
\label{sec:detection}

In this section, we demonstrate the watermark detection capability of our method compared to the baseline approaches.
We also measure the generation quality with watermarks.
Examples of generated images can be seen in Figure~\ref{fig:main_fig_dists} (more in Appendix)

In Figure~\ref{fig:ablations}, we provide the ROC curves with MS-COCO for \name ~with different ablations of top-$k$ or the number of concepts sampled from the database $\mathcal{D}$.
As shown in the plot, the detection performance of \name ~improves as the number of sampled concepts increases.
For instance, the AUROC  and T@1\%F values rise over 21\% and 71\%, respectively, as $k$ changes from 1 to 9 for the MS-COCO captions.
Figure~\ref{fig:ablations} also shows the frequency histogram of the number of concepts detected in the \name ~($k=9$) watermarked and non-watermarked ($k=0$) images.
As shown in the histogram, the watermarked images have a much higher likelihood of having concepts from our concept database than the non-watermarked images.

We also measure the detection and image generation metrics for \name ~for various values of $k$ and provide them in Tables~\ref{tab:detection_metrics} and \ref{tab:generation_metrics}, respectively.
As shown in the tables, the quality of the watermarked images does not degrade with different ablations of $k$, although it shows an increasing trend for the quality of images as $k$ increases. 
In the rest of the analysis, we fix $k=9$ for \name ~since that gives the best detection results without a degradation in the image quality.

\subsection{Robustness of Watermarks}
\label{sec:robustness}

In this section, we evaluate the robustness of our watermarks in the presence of various image manipulation techniques \citep{wam}.
We employ three types of composite modifications to evaluate watermarks' robustness. 
Each composite modification, as shown below, comprises one or more elementary perturbations designed to emulate typical real‐world degradations:

\noindent\textbf{Affine.} Random rotation (in range $[-20^\circ, 20^\circ]$) and random cropping retaining 70\%--95\% of the original area.

\noindent\textbf{Valuemetric.} Photometric distortions, including brightness and contrast adjustments (factors in $[1.4, 1.7]$), Gaussian blur (radius between 1 and 3 pixels), additive Gaussian noise (standard deviation in $[0.05, 0.15]$), and JPEG compression (quality in the range $[40, 70]$).

\noindent\textbf{Regen.} Image regeneration \citep{saberi2023robustness, waves} using a diffusion model with 300 diffusion steps.

\noindent\textbf{Warp.} Perspective warping with image corner locations randomly perturbed within a range of 0\%-40\%.

For examples of these types of augmented images, see Appendix.
We display our results to demonstrate the robustness of our watermarks in Table~\ref{tab:augmentation_comparison} and the corresponding ROC curves in Figure~\ref{fig:robustness}.
As shown in the results, our method \name+SS is the clear winner regarding the detection performance, followed by \name+TM.
\name+SS gets its strength from combining the resilience of \name ~to affine and warp augmentations, and the resilience of StegaStamp to valuemetric augmentations.
Our results show that \name{} and its variants are the only techniques that maintain high detection in the presence of all of the augmentation attacks.
Therefore, on an average \name, \name+SS, and \name+TM achieve 10.8\%, 14.5\%, and 15.9\% higher AUROC when compared to the strongest baseline StegaStamp on both the datasets.

%% file: 5_conclusion.tex
\section{Future Work and Limitations}

By embedding interpretable semantic concepts, \name{} enables watermark detection by both humans and machines for the first time, making it a reliable tool for image forensics and authentication.
Integrating \name ~with existing techniques like StegaStamp and TrustMark further enhances its resilience, making it the most robust method compared to baselines.

\name{} is a first step toward interpretable AI watermarking.
As a proof of concept, it has some limitations.
For instance, \name{} may not fully meet the needs of users with highly specific image generation prompts.
This challenge aligns with theoretical findings on the limitations of AI-generated content detection in low-entropy output spaces, as discussed in \cite{sadasivan2023can} and \cite{saberi2023robustness}. 
In the future, we hope to see variants of \name{} that embed more subtle concepts into the main objects specified by the user, rather than introducing a large number of entirely new objects as watermarks.
As AI models continue to evolve, generation capabilities will expand, and \name ~is expected to become even more effective.
Nevertheless, we believe our novel approach to interpretable watermarks opens up an exciting new avenue for research in this field.

%% file: 6_app.tex
\clearpage
\appendix
\section{Appendix}

\subsection{Additional Examples of Generated Images}

\begin{figure*}[h!]
    \centering
    \begin{subfigure}{0.95\linewidth}
        \centering
        \includegraphics[width=\linewidth]{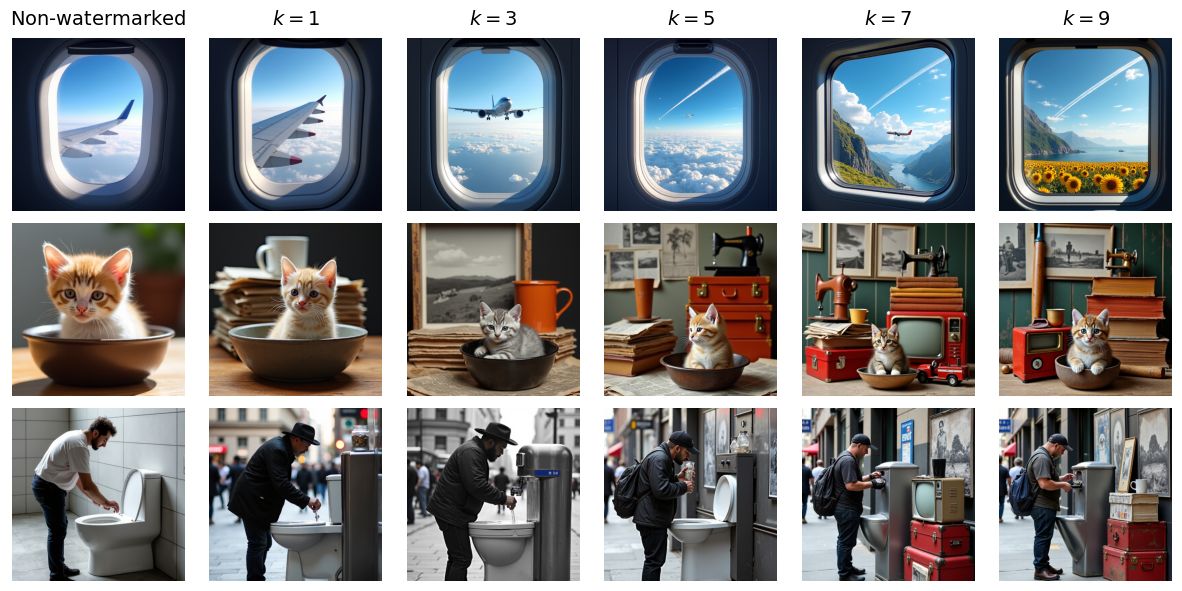}
        \caption{\name ~generation for various numbers of selected concepts $k$.} \label{fig:iconmark}
    \end{subfigure}
    \begin{subfigure}{0.95\linewidth}
        \centering
        \includegraphics[width=\linewidth]{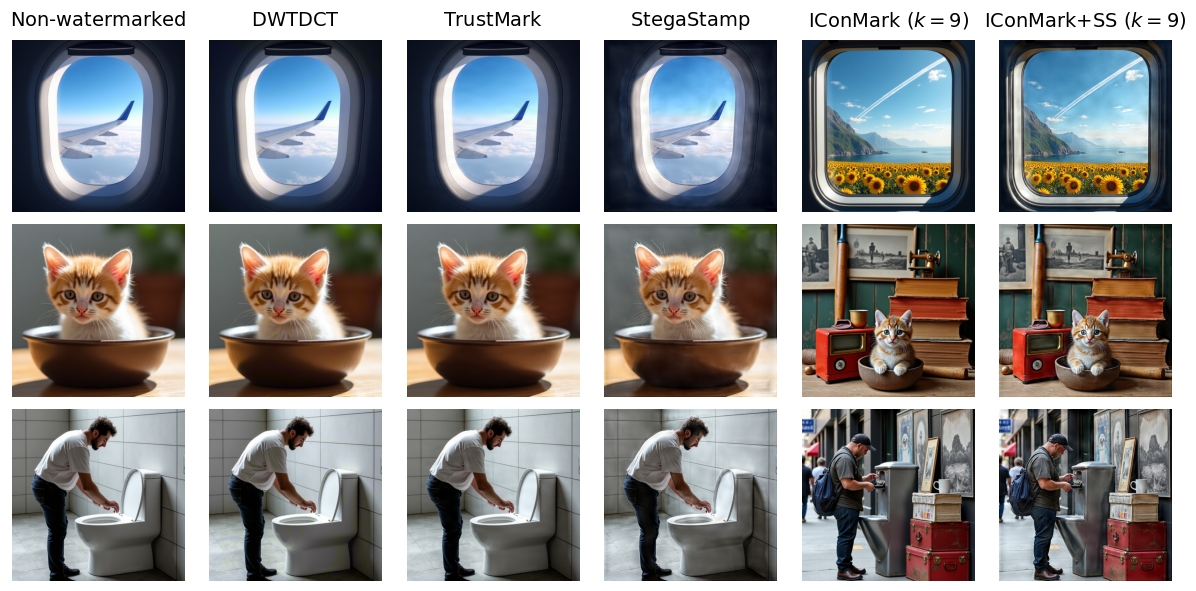}
        \caption{Comparing images with various watermarking techniques.} \label{fig:baselines}
    \end{subfigure}
\caption{Comparing images generated with different watermarking techniques. The images in the first column are non-watermarked AI-generated images from the Flux model. 
For each of the rows of the images, the main user prompts $p$ for generation are ``A view from a window on board an airplane flying in the sky.'', ``A small kitten is sitting in a bowl.'', and ``A man getting a drink from a water fountain that is a toilet.'', respectively. 
In Figure~\ref{fig:iconmark}, we show different ablations of \name ~over the number of sampled concepts from the concept database.
Figure~\ref{fig:baselines} shows the comparison of our methods \name ~and \name+SS with baseline techniques such as DWTDCT \citep{dwtdct}, TrustMark \citep{trustmark}, and StegaStamp \citep{stegastamp}.}
\label{figapp_main}
\end{figure*}

\begin{figure*}[h!]
    \centering
    \includegraphics[width=0.95\linewidth]{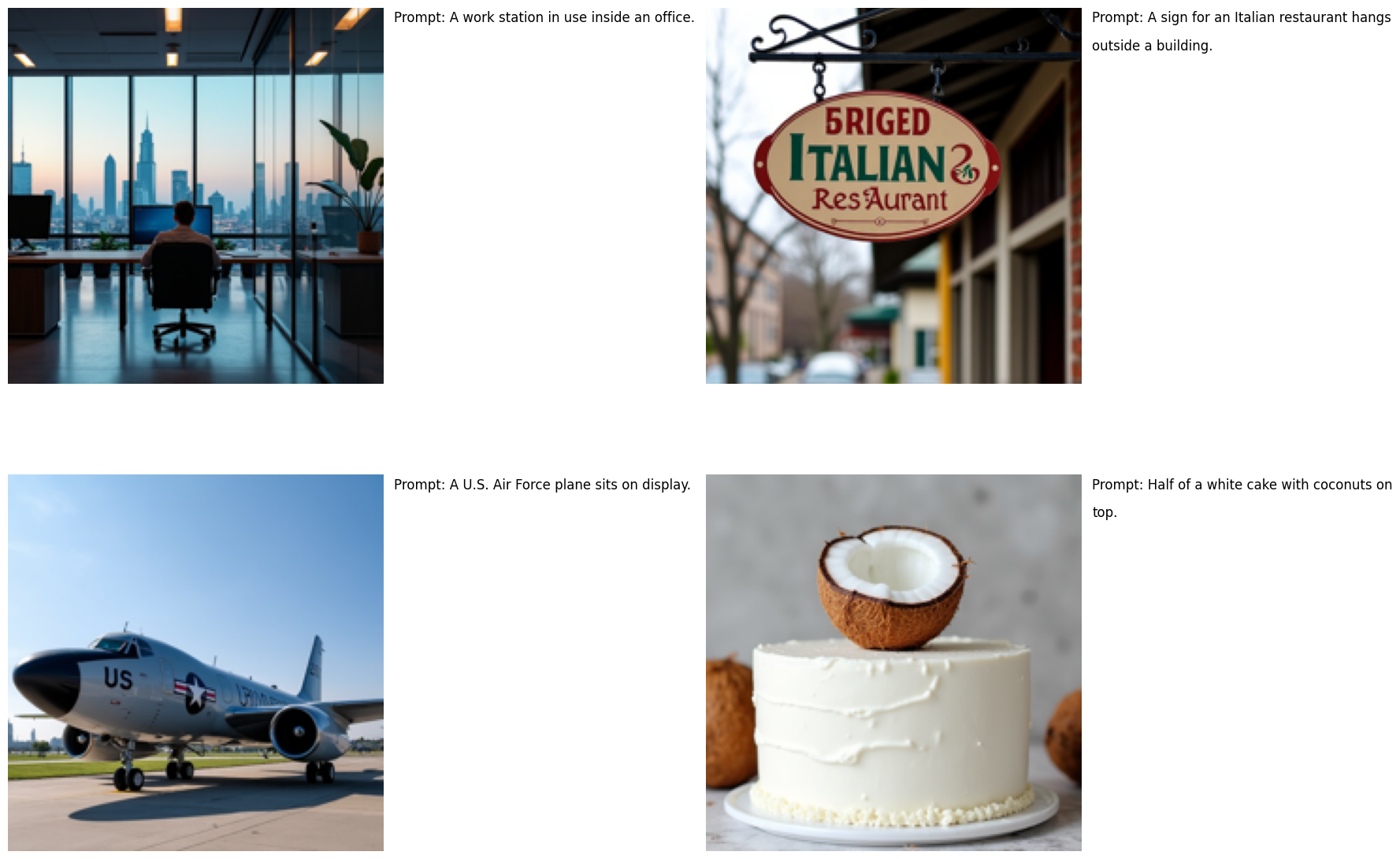}
    \caption{Non-watermarked images with their corresponding prompts for image generation using the Flux model.}
    \label{fig:app-k=0}
\end{figure*}

\begin{figure*}[h!]
    \centering
    \includegraphics[width=0.95\linewidth]{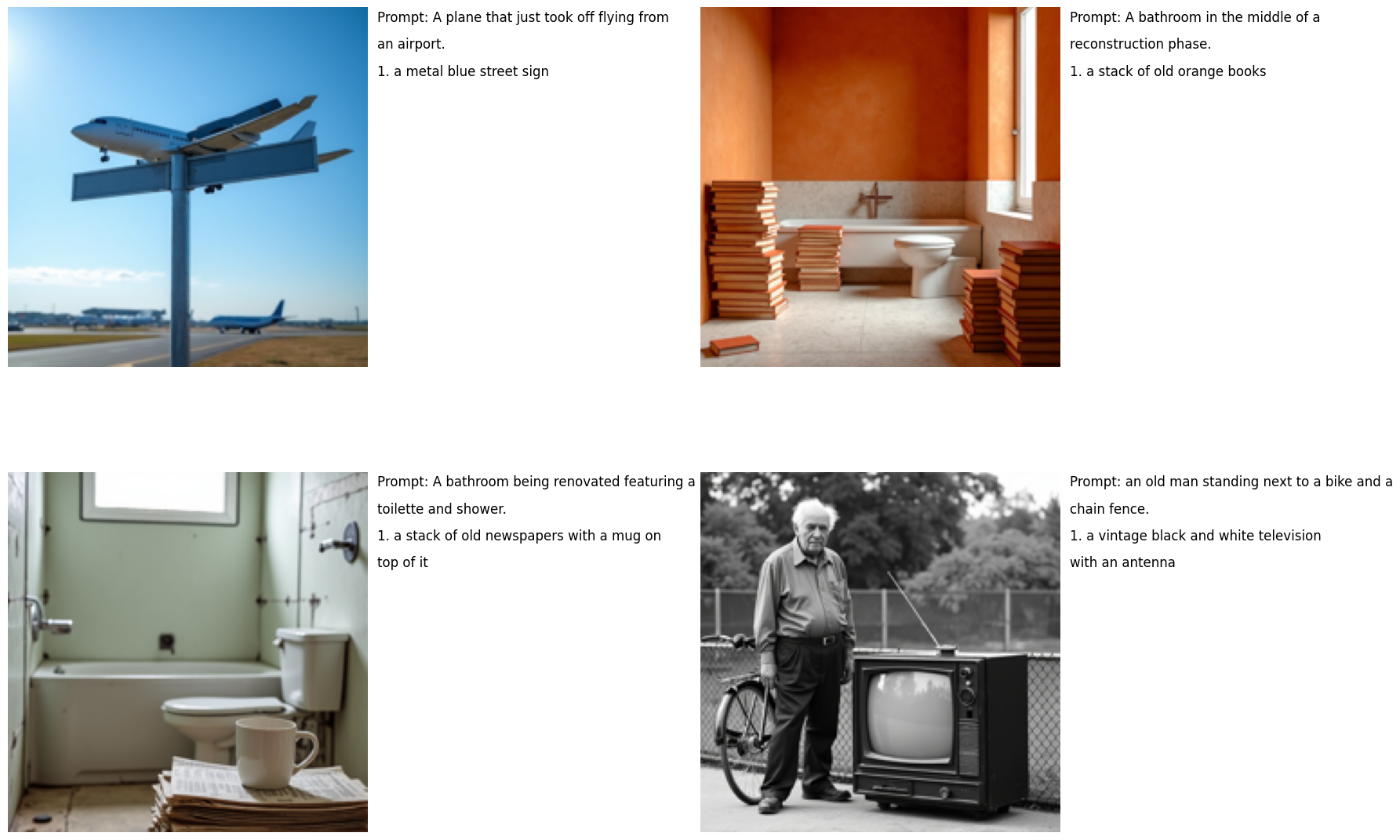}
    \caption{\name ~watermarked images ($k=1$) with their corresponding prompts for image generation using the Flux model and detected concepts from the concept database $\mathcal{D}$ using the IDEFICS3 visual language model.}
    \label{fig:app-k=1}
\end{figure*}

\begin{figure*}[h!]
    \centering
    \includegraphics[width=0.95\linewidth]{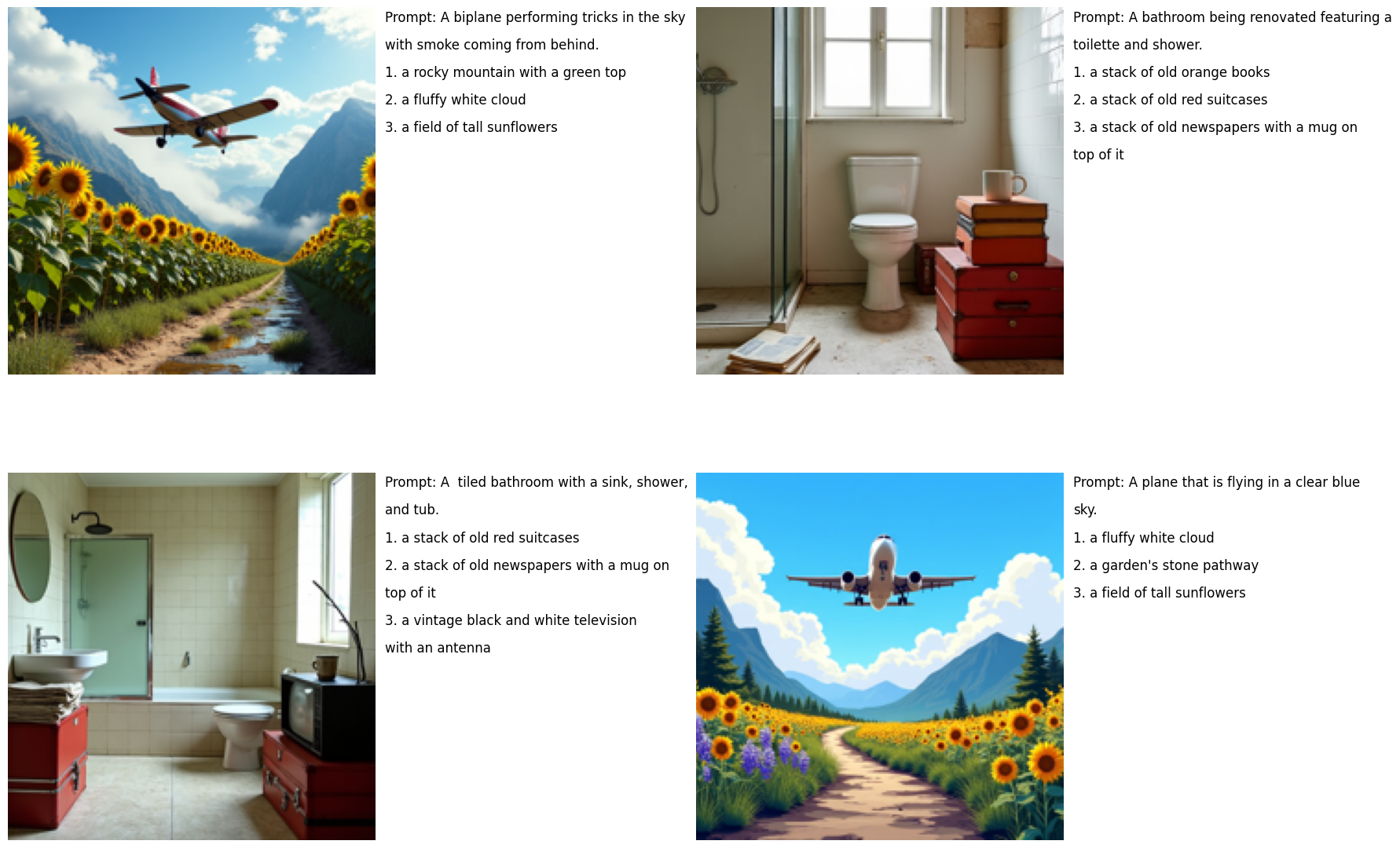}
    \caption{\name ~watermarked images ($k=3$) with their corresponding prompts for image generation using the Flux model and detected concepts from the concept database $\mathcal{D}$ using the IDEFICS3 visual language model.}
    \label{fig:app-k=3}
\end{figure*}

\begin{figure*}[h!]
    \centering
    \includegraphics[width=0.95\linewidth]{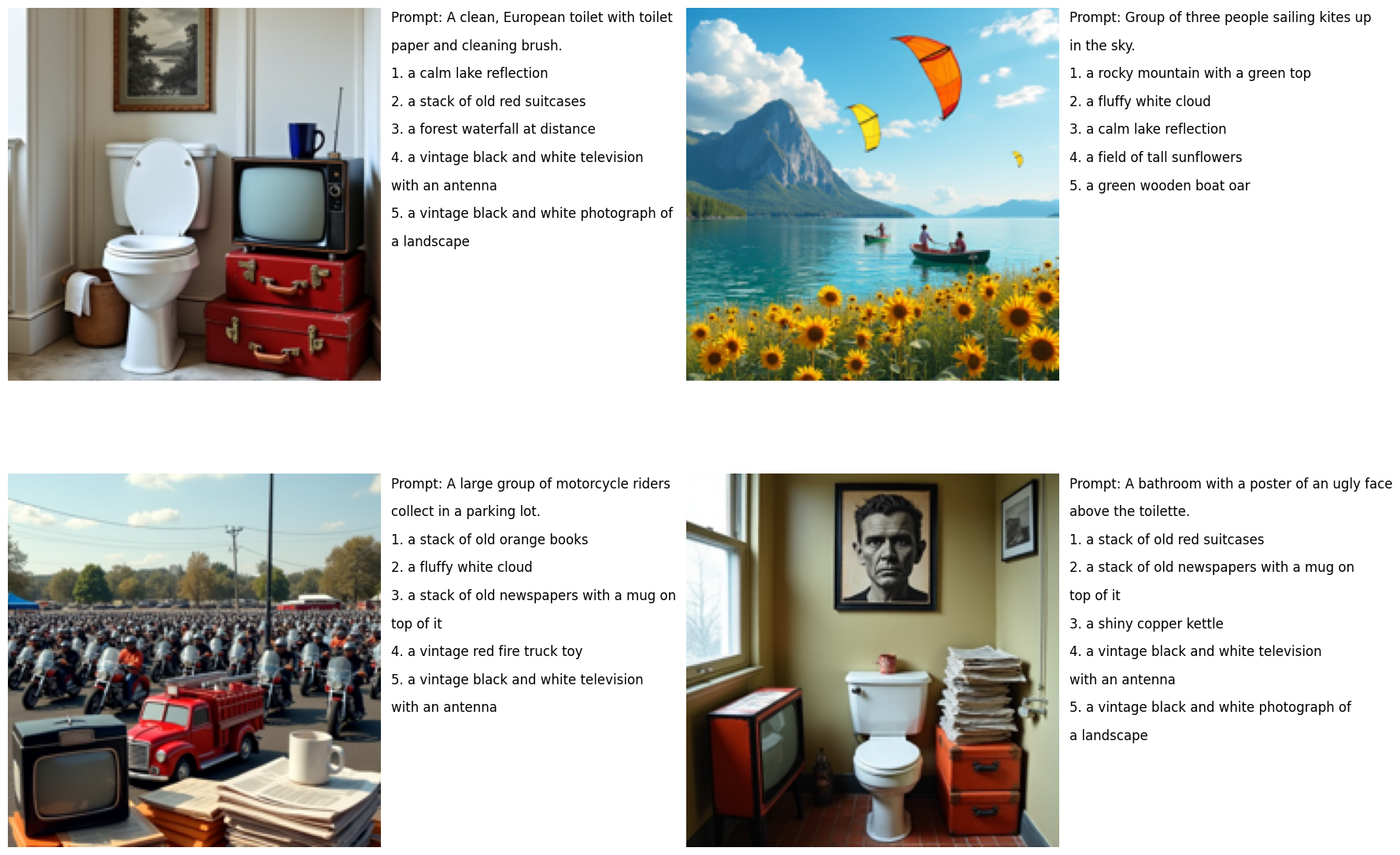}
    \caption{\name ~watermarked images ($k=5$) with their corresponding prompts for image generation using the Flux model and detected concepts from the concept database $\mathcal{D}$ using the IDEFICS3 visual language model.}
    \label{fig:app-k=5}
\end{figure*}

\begin{figure*}[h!]
    \centering
    \includegraphics[width=0.95\linewidth]{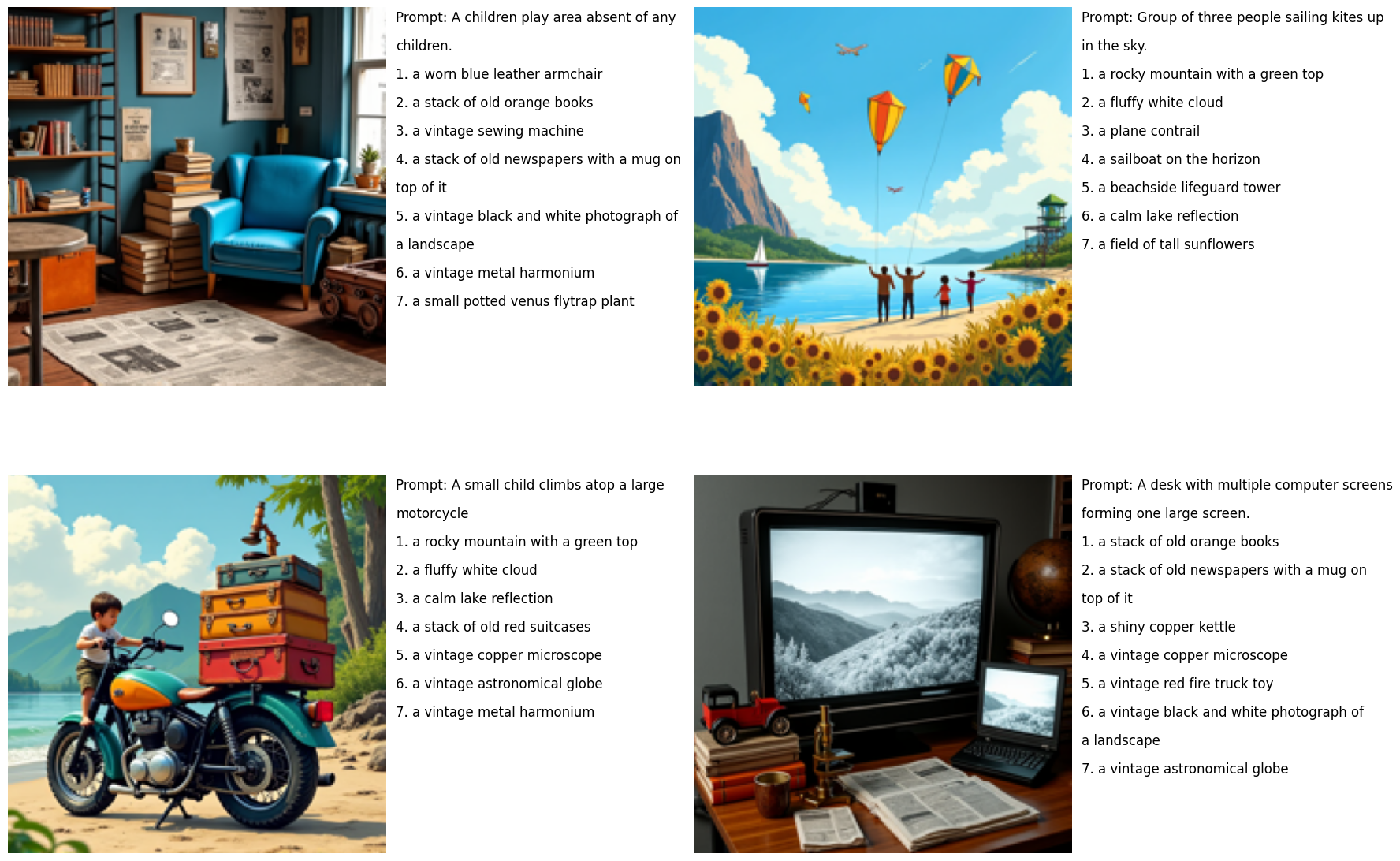}
    \caption{\name ~watermarked images ($k=7$) with their corresponding prompts for image generation using the Flux model and detected concepts from the concept database $\mathcal{D}$ using the IDEFICS3 visual language model.}
    \label{fig:app-k=7}
\end{figure*}

\begin{figure*}[h!]
    \centering
    \includegraphics[width=0.95\linewidth]{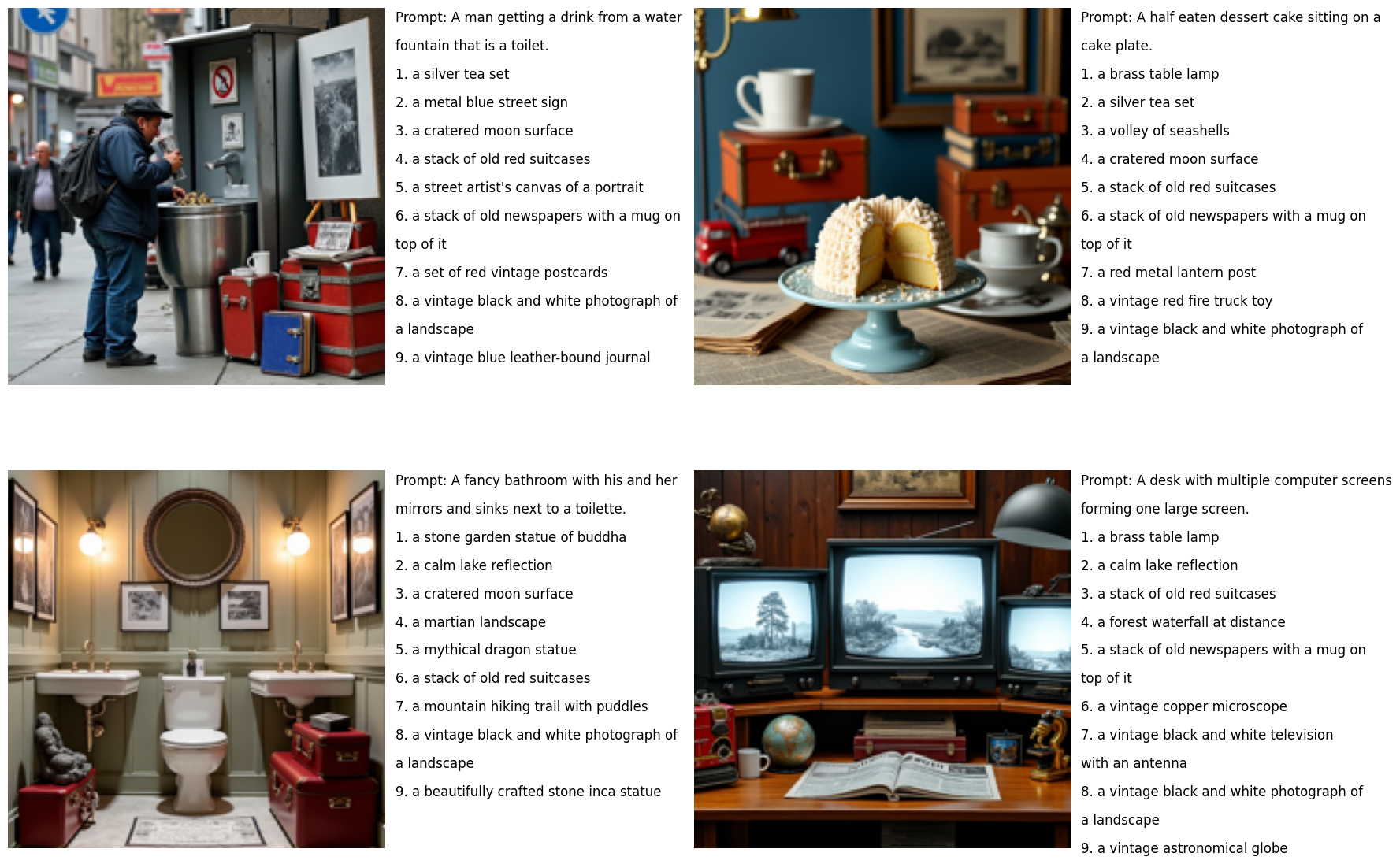}
    \caption{\name ~watermarked images ($k=9$) with their corresponding prompts for image generation using the Flux model and detected concepts from the concept database $\mathcal{D}$ using the IDEFICS3 visual language model.}
    \label{fig:app-k=9}
\end{figure*}

\begin{figure*}[h!]
    \centering
    \includegraphics[width=\linewidth]{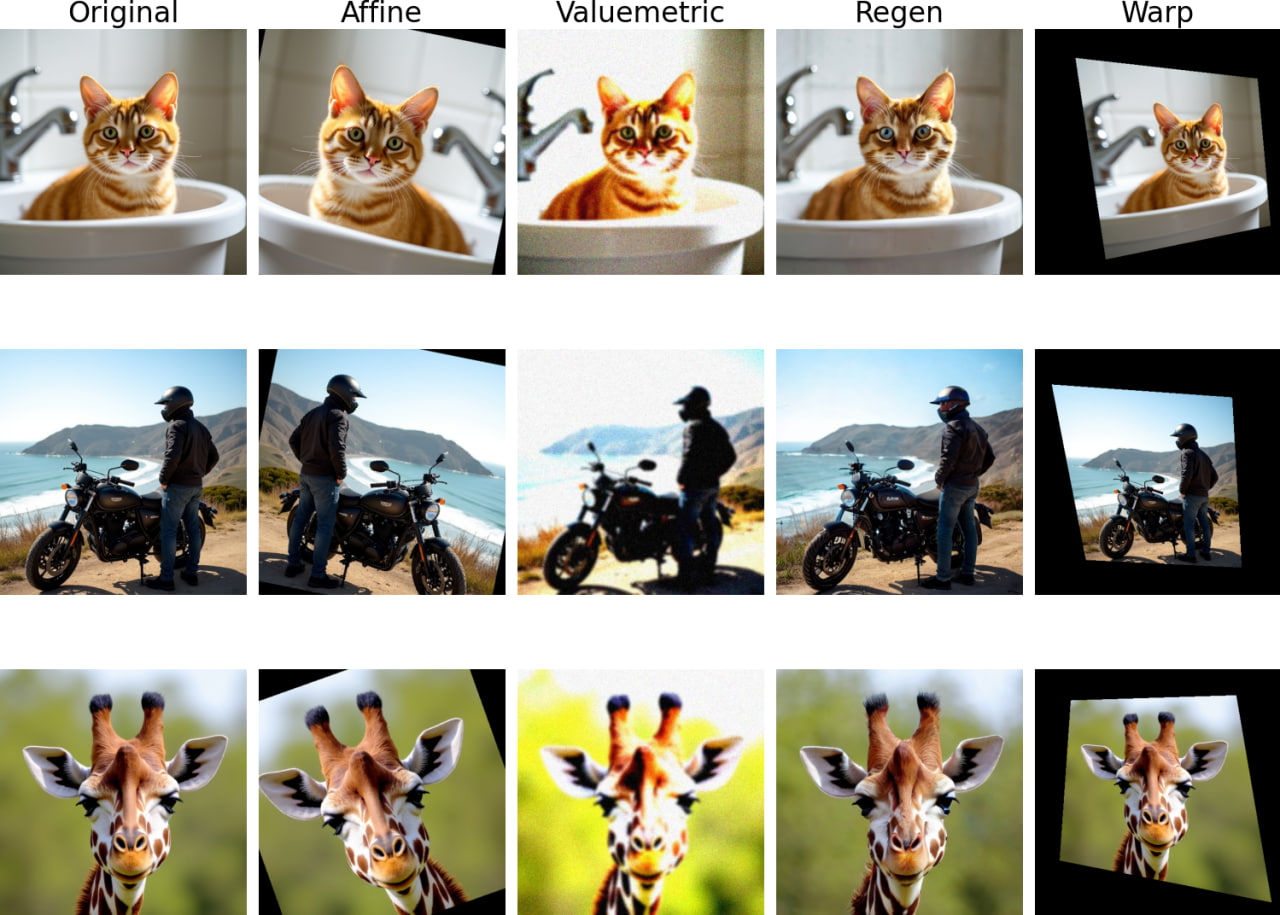}
    \caption{Examples of modified images for each modification type used in Section~\ref{sec:robustness}.}
    \label{fig:aug_examples}
\end{figure*}

\begin{tcolorbox}[breakable, enhanced, title=Concept Database]
\small
1. a worn blue leather armchair \\
2. a brass table lamp \\
3. a stack of old orange books \\
4. a white vintage typewriter \\
5. a red brick fireplace \\
6. a patterned green persian rug \\
7. a silver tea set \\
8. a glass chandelier \\
9. a red wooden grandfather clock \\
10. a collection of vinyl records in blue \\
11. a stone garden statue of buddha \\
12. a bright pink colored beach umbrella \\
13. a metal blue street sign \\
14. a yellow public phone booth \\
15. a smooth river rock next to a pebble \\
16. a patch of colorful purple wildflowers \\
17. a heap of fallen autumn leaves \\
18. a bird's nest with eggs \\
19. a moss-covered tree trunk with a hole \\
20. a rocky mountain with a green top \\
21. a fluffy white cloud \\
22. a crescent moon shape with two stars next to it \\
23. a vague rainbow arc \\
24. a plane contrail \\
25. a yellow hot air balloon \\
26. a sailboat on the horizon \\
27. a piece of floating green seaweed \\
28. a rusted ship's anchor \\
29. a beachside lifeguard tower \\
30. a calm lake reflection \\
31. a volley of seashells \\
32. a distant planet's red ring \\
33. a cratered moon surface \\
34. a rocket ship's engines \\
35. a martian landscape \\
36. a comet's tail with two stars next to it \\
37. a magical yellow crystal ball \\
38. a mythical dragon statue \\
39. a fantasy castle tower \\
40. a futuristic robot arm \\
41. a mythical unicorn horn \\
42. a vintage sewing machine \\
43. a blue wooden picture frame \\
44. a beautifully crafted black music box \\
45. a stack of old red suitcases \\
46. a bright yellow colored food cart \\
47. a city street performer's tip jar with coins \\
48. a park's walking trail sign in green color \\
49. a garden's stone pathway \\
50. a green beach volleyball \\
51. a parking meter in black color \\
52. a street artist's canvas of a portrait \\
53. a beehive in a tree \\
54. a forest waterfall at distance \\
55. a field of tall sunflowers \\
56. a desert cactus spine with blue spots \\
57. a mountain hiking trail with puddles \\
58. a red fire extinguisher \\
59. a green wooden boat oar \\
60. a stack of old newspapers with a mug on top of it \\
61. a beautifully crafted brown wooden flute \\
62. a small potted cactus with red spots \\
63. a woven basket with blueberries \\
64. a red metal lantern post \\
65. a set of orange gardening gloves \\
66. a beautifully crafted blue wooden birdhouse \\
67. a circular metal street grate \\
68. a beautifully crafted stone fountain \\
69. a beautifully crafted yellow wooden rocking chair \\
70. a medieval castle wall with ferns on it \\
71. a dark and spooky cave with dead trees around it \\
72. a wooden treasure chest with gold \\
73. a metal astronaut's helmet \\
74. a metal submarine's propeller \\
75. a red colored party hat \\
76. a shiny copper kettle \\
77. a beautifully crafted wooden carousel horse \\
78. a vintage copper microscope \\
79. a worn wooden baseball bat with a blue grip \\
80. a bright orange construction cone \\
81. a small potted bonsai tree with pink and blue flowers \\
82. a vintage red fire truck toy \\
83. a beautifully crafted wooden model ship in blue color \\
84. a set of red vintage postcards \\
85. a vintage black and white television with an antenna \\
86. a set of antique binoculars in brown color \\
87. a beautifully crafted circular wooden wall mirror \\
88. a beautifully crafted crystal decanter half-filled with wine \\
89. a vintage black and white photograph of a landscape \\
90. a brown wooden walking stick with a silver handle \\
91. a set of fine silver picture frames with engravings \\
92. a vintage astronomical globe \\
93. a beautifully crafted wooden abacus \\
94. a vintage metal harmonium \\
95. a small potted venus flytrap plant \\
96. a vintage blue leather-bound journal \\
97. a beautifully crafted wooden model of the eiffel tower \\
98. a set of blue and white striped candy canes \\
99. a beautifully crafted stone inca statue \\
100. a red acoustic guitar \\
\end{tcolorbox}